\documentclass[runningheads]{llncs}
\usepackage[T1]{fontenc}
\usepackage{booktabs}
\usepackage{array}

\usepackage{graphicx}
\usepackage{color}
\usepackage{amsmath,amsfonts,amssymb}
\usepackage{adjustbox}
\usepackage{rotating}
\usepackage{orcidlink}
\usepackage{esvect}

\begin{document}
\title{A Fast and Efficient Modern BERT based Text-Conditioned Diffusion Model for Medical Image Segmentation}
\titlerunning{FastTextDiff for Medical Image Segmentation}

%
\author{Venkata Siddharth Dhara\inst{1}\orcidlink{0009-0000-3036-0785} \and
Pawan Kumar\inst{2}\orcidlink{0000-0001-5632-6964}}
\authorrunning{V. S. Dhara and P. Kumar}
%
\institute{International Institute of Information Technology, Hyderabad, 500032, India\\
\email{venkatasiddharth.d@research.iiit.ac.in, pawan.kumar@iiit.ac.in}}
\maketitle              
\begin{abstract}
In recent times, denoising diffusion probabilistic models 
(DPMs) have proven to show significant success in medical image generation and denoising, while also serving as powerful representation learners for downstream tasks such as segmentation. However, their effectiveness in segmentation is limited by the need for detailed pixel-wise annotations, which are expensive, time-consuming, and require expert knowledge—a significant bottleneck in real-world clinical applications. In order to mitigate this limitation of label-efficiency, we propose a fast and efficient model named FastTextDiff, a diffusion-based segmentation model that integrates medical text annotations to enhance semantic representations. Our approach leverages ModernBERT \cite{modernbert}, a transformer-based language model capable of processing long medical text sequences, to establish a strong connection between textual annotations and semantic meaning in medical imaging. ModernBERT can efficiently encode clinical knowledge for directing segmentation tasks since it has been trained on both MIMIC-III \cite{johnson2016mimic} and MIMIC-IV. Label-efficient segmentation with enhanced performance is made possible by cross-modal attention processes, which enable smooth interaction between visual and textual modalities. This study validates ModernBERT as a quick and scalable substitute for Clinical BioBERT \cite{alsentzer2019publicly} in diffusion-based segmentation pipelines \cite{feng2024enhancing} and demonstrates the promise of multi-modal techniques for medical image analysis. By replacing Clinical BioBERT with ModernBERT, our model benefits from Flash Attention 2 for memory-efficient training, an alternating attention mechanism for computational efficiency, and a training corpus of 2 trillion tokens, significantly improving text-based medical image segmentation. Our experiments demonstrate that FastTextDiff achieves better segmentation performance and faster training compared to traditional diffusion-based models. 

Code is publicly available at:
\url{https://github.com/siddharthdhara/FastTextDiff}

\end{abstract}

\section{Introduction}

Medical image segmentation is a key component of several clinical applications, which includes disease diagnosis, treatment planning etc, wherein we define the regions of interest (ROIs) or boundaries in medical images. Over the past few years, deep learning models, particularly CNNs and its variants like U-Net \cite{ronneberger2015u} and transformer-based models \cite{chen2021transunet,cao2021swin}, have been used for segmentation, and do provide good segmentation accuracy. However, all these methods are completely reliant on extensive datasets (data augmentation is sometimes used \cite{Yang2022,zhao2019,Sandfort2019,Steiner2022,sri2024}) with precise, pixel-level annotations which are manual, costly, time-consuming and there is a need for domain-specific knowledge from clinicians like radiologists or pathologists to produce such annotations \cite{feng2017discriminative,xu2019camel}. This proves to be a major bottleneck, which further affects the scalability and practical implementation of such segmentation models in clinical settings. We thus show significant gains in inference and training speed compared to the TextDiff Architecture \cite{feng2024enhancing} which uses a less optimized text encoder.

Concurrently, Denoising Diffusion Probabilistic Models (DPMs) \cite{ho2020denoising,dhariwal2021diffusion} have shown to be powerful generative models, with state-of-the-art performance across vision related tasks \cite{croitoru2022diffusion}. Their strong generative capabilities suggest significant potential in medical imaging applications, including image synthesis \cite{peng2022generating,kim2022diffusion,moghadam2023morphology}, denoising, and serving as feature extractors for downstream tasks \cite{kazerouni2022diffusion}. Recent work also indicates that there is also a possibility of re-purposing these for segmentation \cite{baranchuk2021label}, wherein diffusion models can learn rich and detailed representations \cite{feng2023diverse}. However, as with other supervised approaches, fine-tuning of these models for segmentation is still often reliant on heavy pixel-wise labels.

To address this problem of label efficiency, it is a good practice to leverage supplementary information that is readily available in clinical practice, such as textual diagnostic reports or annotations as a compelling alternative. These textual descriptions tend to contain rich semantic information which could potentially guide the segmentation process, thus reducing the reliance of heavy pixel level maps. Inspired by this idea, we present \textbf{FastTextDiff}, a novel text-conditioned medical image segmentation framework based on diffusion models, extending the concepts presented in TextDiff \cite{feng2024enhancing}.

Our primary contribution is the incorporation of \textbf{ModernBERT} \cite{modernbert}, a powerful and efficient transformer-based language model, as the text encoder in a diffusion-based segmentation pipeline. Because ModernBERT has been pre-trained on big corpora, which we then retrain on large clinical corpora (MIMIC-III \cite{johnson2016mimic} and MIMIC-IV), it can analyze lengthy clinical text sequences effectively. Additionally, it features alternating attention methods and architectural improvements like Flash Attention 2. Using cross-modal attention, FastTextDiff combines the textual embeddings from ModernBERT with the visual characteristics from the diffusion model's U-Net backbone to effectively translate textual descriptions into spatial information for segmentation. These improvements underscore the practical value of FastTextDiff in resource-constrained clinical settings and its potential for scalable deployment.

The following are the main contributions of our work:
\begin{itemize}
 \item We provide FastTextDiff, a text-conditioned diffusion model for medical picture segmentation that is fast and efficient, using ModernBERT \cite{modernbert} for text encoding.
 \item Thanks to large-scale pre-training and transformer improvements, we verify ModernBERT as a computationally efficient and highly effective substitute for models like Clinical BioBERT \cite{alsentzer2019publicly} within this framework.
 \item Through studies on multiple datasets (MoNuSeg \cite{kumar2019multi}, QaTa-COV19 \cite{degerli2022osegnet}, and MosMedData+ \cite{li2021dual}), we demonstrate that FastTextDiff delivers competitive or state-of-the-art segmentation performance when compared to baseline diffusion models and other vision-language approaches.
 \item We show significant gains in training and inference performance compared to a comparable architecture with a less optimized text encoder (TextDiff baseline \cite{feng2024enhancing}).
\end{itemize}

\section{Related Work}
Medical image segmentation, diffusion models in medical imaging, vision-language models, and clinical text processing are some of the research areas that our work touches upon.

\noindent\textbf{Medical Image Segmentation:} Deep learning has revolutionized medical image segmentation. U-Net \cite{ronneberger2015u} and its numerous variants remain strong baselines due to their effectiveness in capturing multi-scale spatial context. Though sometimes at the expense of more parameters and computational requirements, Vision Transformers (ViTs) and hybrid architectures such as TransUNet \cite{chen2021transunet} and SwinUNet \cite{cao2021swin} have demonstrated potential in recent years by using self-attention to represent long-range dependencies. The requirement for huge annotated datasets is still a major limitation of all these approaches \cite{yu2019uncertainty}.

\noindent\textbf{Diffusion Models in Medical Imaging:} Beyond generation, DPMs are becoming more and more popular in medical imaging \cite{kazerouni2022diffusion}. They have been applied to the synthesis of realistic medical images \cite{peng2022generating}, anomaly detection, and image denoising \cite{kim2022diffusion}. Using the intermediate representations that DPMs learn for downstream tasks like segmentation \cite{baranchuk2021label} and classification has been investigated in a number of research studies \cite{feng2023diverse}. Despite their strength, these models usually require fine-tuning, which might still be data-hungry, in order to be used for supervised segmentation. While some studies use diffusion models to investigate unsupervised or weakly supervised segmentation \cite{feng2021task}, there is still work being done to incorporate explicit semantic guidance.

\noindent\textbf{VLMs (Vision-Language Models):} There is a lot of promise for medical AI in combining textual and visual data. Medical VLMs, which draw inspiration from general-domain VLMs like as CLIP and ViLBERT, seek to learn joint representations from pictures (such as pathology slides, CT scans, and X-rays) and related text (such as radiology reports and clinical notes). Medical visual question answering, report creation, and picture classification/retrieval are among the tasks in which models such as GLoRIA \cite{huang2021gloria}, LViT \cite{li2022lvit}, and others have shown success. Some may not specifically make use of the generative capability or structured feature space of diffusion models, but have been modified for segmentation, often by trying to align different regions of the image with text descriptions.

\noindent\textbf{Text-Conditioned Segmentation:} A rather new/emerging method involves conditioning segmentation or generative models on text. Text prompts have typically been used in the general domain to create images \cite{dhariwal2021diffusion}, or to alter images \cite{ruiz2022dreambooth}. Applying this to medical segmentation allows leveraging clinical descriptions. The TextDiff approach \cite{feng2024enhancing}, upon which our work builds, demonstrated the feasibility of using a text encoder (like Clinical BioBERT \cite{alsentzer2019publicly}) combined with a diffusion model for segmentation. This is furthered in our work by integrating a more modern and effective text encoder.

\noindent\textbf{BERT Variants in Medical NLP:} Processing clinical text requires specialized language models. Models like BioBERT and ClinicalBERT \cite{alsentzer2019publicly}, which are frequently pre-trained on MIMIC data \cite{johnson2016mimic} or PubMed abstracts, are the outcome of BERT's adaptation for the biomedical and clinical domains. Although these models are good at capturing clinical semantics, they may not be as efficient when dealing with lengthy sequences or large-scale deployment. ModernBERT \cite{modernbert} is a recent development that is well-suited for computationally demanding tasks like integration into diffusion pipelines because it was trained on very large corpora, and was built for efficiency using techniques like Flash Attention 2 and alternating attention. As far as we are aware, our work is the first, to evaluate and demonstrate the benefits of ModernBERT within a text-conditioned diffusion model for medical image segmentation.

\section{Methodology: FastTextDiff}

\subsection{Background: Diffusion Models}

Diffusion models \cite{ho2020denoising} are generative models that gradually corrupt an image with noise (forward process) and then learn to reverse the process (backward process) using a learned denoiser.

\subsubsection{Forward Process}
The forward diffusion process involves adding of Gaussian noise gradually to a clean input image \( x_0 \) over \( T \) timesteps. At each timestep \( t \), the noisy image \( x_t \) is generated as:
\begin{equation}
    q(x_t \mid x_{t-1}) = \mathcal{N}\Big(x_t; \sqrt{1-\beta_t}\,x_{t-1},\,\beta_t \mathbf{I}\Big),
    \label{eq:forward}
\end{equation}
where \( \beta_t \) is a variance schedule that controls the noise level at each timestep. This process is well-defined and allows sampling \( x_t \) directly from \( x_0 \).

\subsubsection{Backward Process}
The reverse process involves recovery of the clean input image \( x_0 \) from the noise \( x_T \). This is typically modeled as a Markov chain starting from \( p(x_T) = \mathcal{N}(x_T; \mathbf{0}, \mathbf{I}) \):
\begin{equation}
    p_\theta(x_{t-1} \mid x_t) = \mathcal{N}\Big(x_{t-1}; \mu_\theta(x_t,t),\,\Sigma_\theta(x_t,t)\Big),
    \label{eq:backward}
\end{equation}
where \( \mu_\theta \) and \( \Sigma_\theta \) are predicted by a neural network (the denoiser), parameterized by \( \theta \).

\subsubsection{Denoiser Network}
The denoising network, a U-Net architecture \cite{ronneberger2015u} is trained to predict the noise component \( \epsilon \) added at timestep \( t \), given the noisy image \( x_t \) and the timestep \( t \):
\begin{equation}
    \epsilon_\theta(x_t, t) \approx \epsilon,
    \label{eq:denoiser}
\end{equation}
using a loss function like Mean Squared Error between the the actual noise \( \epsilon \) and the predicted noise \( \epsilon_\theta \). This predicted noise is then used to estimate \( \mu_\theta(x_t,t) \) for the reverse step \eqref{eq:backward}. This forms the backbone of many state-of-the-art diffusion models \cite{dhariwal2021diffusion}.

\subsection{Crossmodal Attention}
Cross-modal attention is an attention mechanism designed to fuse and align features from different modalities (e.g., text and images). It has been widely adopted in vision-language tasks (see, e.g., \cite{huang2021gloria}).

\subsection{Mechanism}
Given image features \( h \) (from intermediate layers of the U-Net) and text features \( \tilde{t} \) (from ModernBERT), the scaled dot-product attention can be expressed as:
\begin{equation}
    H_{z,t} = \text{Softmax}\!\left(\frac{Q K^\top}{\sqrt{d_k}}\right)V,
    \label{eq:cross_attention_general}
\end{equation}
where \( K = \tilde{t} W_k \), \( Q = h W_q \), and \( V = \tilde{t} W_v \) are the key, query, and value matrices derived from the image and text features using learnable projection matrices \( W_q \), \( W_k \), and \( W_v \). \( d_k \) is the dimension of the keys. The output \( H_{z,t} \) represents the enhanced image feature representations, modulated by the textual information, which are then used in subsequent layers of the segmentation model.

\section{FastTextDiff Framework}

\begin{figure}[!htbp]
    \centering
    \includegraphics[width=1.0\linewidth]{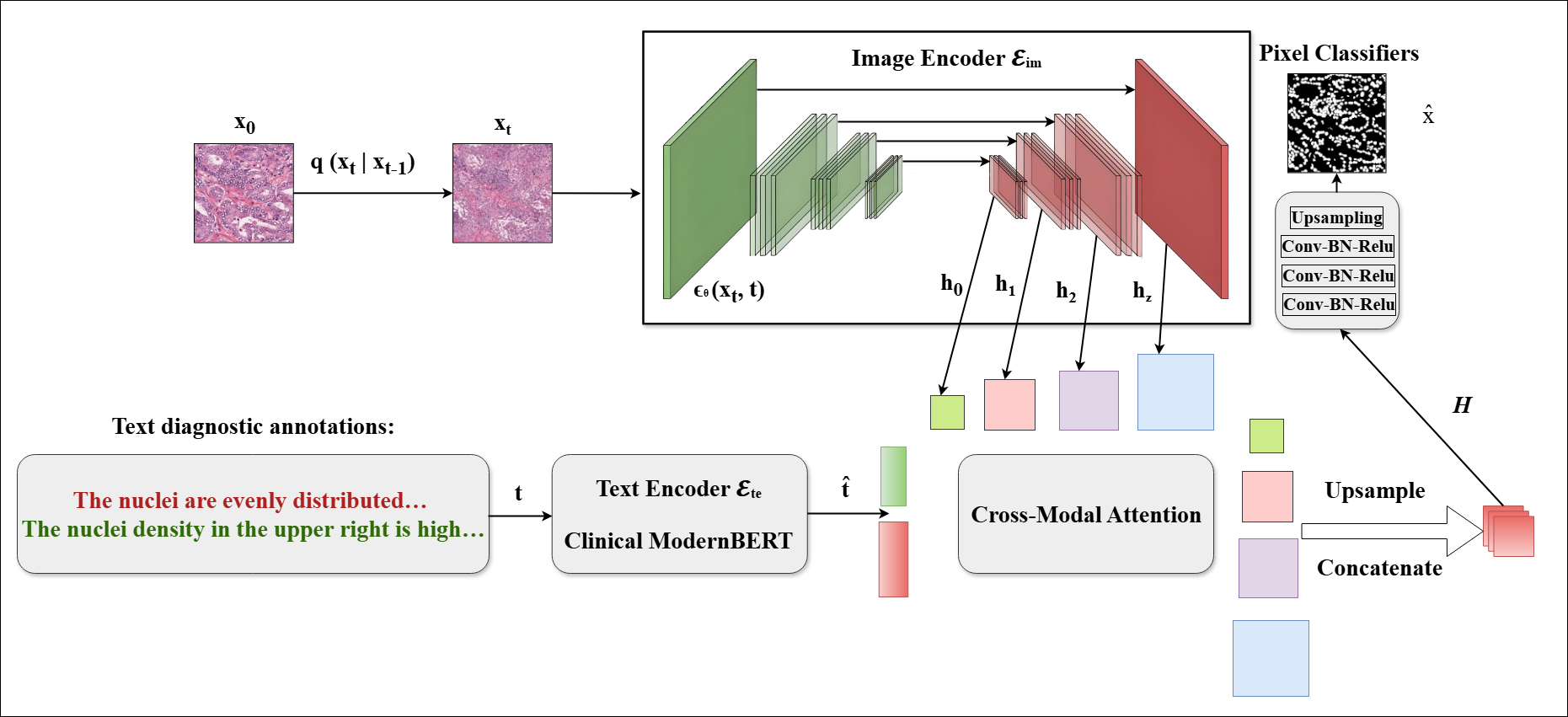}
    \caption{Architecture of our proposed FastTextDiff method, integrating ModernBERT text embeddings with a U-Net based diffusion model via cross-modal attention for text-conditioned segmentation.}
    \label{fig:architecture}
\end{figure}

The proposed FastTextDiff framework integrates text conditioning into a diffusion-based segmentation pipeline for improved performance and label efficiency. It builds upon the TextDiff concept \cite{feng2024enhancing} but incorporates ModernBERT \cite{modernbert} for enhanced text encoding and efficiency. The overall architecture is shown in Figure~\ref{fig:architecture}.

\subsection{Architecture Overview}
FastTextDiff consists of two main branches integrated via cross-modal attention:
\begin{itemize}
    \item \textbf{Image Encoder:} A pre-trained diffusion model (typically based on a U-Net architecture similar to \cite{ho2020denoising}) extracts intermediate visual features \( h \) from noisy versions of the input medical images \( x_t \) at various timesteps \( t \).
    \item \textbf{Text Encoder:} A pre-trained text model, ModernBERT \cite{modernbert}, processes the corresponding diagnostic text description \( t \) to produce contextual text embeddings \( \tilde{t} \). ModernBERT is chosen for its efficiency with long sequences (due to alternating attention and Flash Attention 2) and its extensive pre-training, further fine-tuned here on MIMIC-III \cite{johnson2016mimic} and MIMIC-IV datasets.
\end{itemize}
These branches are integrated via cross-modal attention modules (as described in Equation~\ref{eq:cross_attention_general}) within the U-Net decoder blocks. These modules align and fuse the visual features \( h \) with the text embeddings \( \tilde{t} \). The fused representation guides the denoising process, implicitly incorporating semantic information from the text into the prediction. The output of the final layer, after processing the fused features, is used to produce the segmentation mask \( \hat{x} \). Also, during training only the cross modal block and pixel classifier block are trainable, and rest of the blocks are frozen. 

\subsection{Loss Function}
The training objective of FastTextDiff, following common practice in segmentation, combines the cross-entropy loss and the Dice loss to balance spatial overlap accuracy with pixel-wise classification performance:
\begin{equation}
    L = L_{\text{Dice}} + L_{\text{CE}}.
    \label{eq:loss}
\end{equation}
This loss is computed between the predicted segmentation mask \( \hat{x} \) and the ground truth mask.

\subsection{Inputs and Outputs}
\begin{itemize}
    \item \textbf{Input:} A pair \( (x, t) \) where \( x \) is a medical image and \( t \) is the corresponding diagnostic text. Internally, the model operates on noisy versions \( x_t \).
    \item \textbf{Output:} A segmentation mask \( \hat{x} \) that localizes the region of interest described or relevant to the text \( t \) within the image \( x \).
\end{itemize}

\section{Numerical Experiments}

\subsection{Experimental Setup}
We implement our method using \textbf{PyTorch} and \textbf{Hugging Face Transformers} as the primary libraries. All our experiments are done on a single \textbf{NVIDIA RTX 4090 GPU} with 24~GB of VRAM. \textbf{FastTextDiff} is trained for \textbf{50 epochs} using the \textbf{Adam} optimizer, with an initial learning rate of \texttt{1e-4} and a batch size of 1. All input images are downsized to \textbf{256~$\times$~256}.

For the U-Net decoder in the diffusion backbone, we adopt different middle block configurations depending on the dataset, following the setup in \cite{feng2024enhancing}:
\begin{itemize}
    \item \textbf{MoNuSeg}: $B = \{6, 8, 12, 16\}$ with diffusion time steps $t = \{50, 150, 250\}$
    \item \textbf{QaTa-COVID19} and \textbf{MosMedDataPlus}: $B = \{4, 6, 8, 12\}$ with the same time steps $t = \{50, 150, 250\}$
\end{itemize}

Model performance is evaluated using the \textbf{Intersection over Union (IoU)} and the \textbf{Dice Similarity Coefficient (DSC)} metrics, standard for segmentation tasks.

\subsection{Datasets}

We evaluate the effectiveness of our proposed method on three publicly available medical imaging datasets: \textbf{MoNuSeg} \cite{kumar2019multi}, \textbf{QaTa-COV19} \cite{degerli2022osegnet}, and \textbf{MosMedData+} (derived from MosMedData, often used in COVID studies like \cite{li2021dual}). These datasets span multiple modalities, including pathology and chest CT/X-ray scans, and offer a diverse evaluation of label-efficient segmentation performance under limited supervision.

\begin{itemize}
    \item \textbf{MoNuSeg}: The Multi-Organ Nucleus Segmentation dataset originates from the MICCAI 2018: MoNuSeg challenge \cite{kumar2019multi}. It consists of 30 pathology images with 21,623 annotated nuclear boundaries for training, and the remaining 14 images with 7,000 annotated boundaries for testing. We utilize the complete dataset for training and evaluation, although the text conditioning relies on generic class descriptions rather than per-image reports for this dataset.
    \item \textbf{QaTa-COV19}: This COVID-19 chest radiography dataset \cite{degerli2022osegnet} contains chest X-ray images with pixel-wise lesion annotations and associated textual diagnostic reports. For our experiments, we follow the split described in \cite{feng2024enhancing}: randomly sampling 150 training images, 23 validation images, and 34 test images (a 70-10-20 split), and use the accompanying text annotations to enhance vision-language fusion during segmentation.
    \item \textbf{MosMedDataPlus}: This is an extended version of the MosMedData dataset (related work often uses it for COVID segmentation, e.g., \cite{li2021dual}), containing CT scans. For efficient training under resource constraints, we follow the split from \cite{feng2024enhancing}: randomly select 400 images from the training set, 50 from the validation set, and 50 from the test set, following an approximate 80-10-10 split. Text descriptions (likely related to COVID findings) are used for conditioning.
\end{itemize}

The final data distribution used in our experiments is summarized in Table~\ref{tab:data_distribution}.

\begin{table}[!htbp]
\centering
\caption{Dataset splits used in our experiments (following \cite{feng2024enhancing}).}
\label{tab:data_distribution}
\begin{tabular}{|l|c|c|c|c|}
\hline
\textbf{Dataset} & \textbf{Train Set} & \textbf{Validation Set} & \textbf{Test Set} & \textbf{Total Used} \\
\hline
MoNuSeg \cite{kumar2019multi}         & 24                & 6                       & 14                & 44             \\
QaTa-COV19 \cite{degerli2022osegnet}      & 150               & 23                      & 34                & 207            \\
MosMedData+ \cite{li2021dual}     & 400               & 50                      & 50                & 500            \\
\hline
\end{tabular}
\end{table}

\subsection{Experimental Results and Performance Comparison}

We compare FastTextDiff against the original TextDiff \cite{feng2024enhancing} and other state-of-the-art segmentation methods. Table~\ref{tab:results} shows the impact of different text encoders and pre-training on segmentation performance. ModernBERT consistently outperforms the baseline BERT used in the original TextDiff paper, with MIMIC pre-training providing a slight edge, especially with longer context lengths (512 vs 128) on certain datasets like QaTa-COV19 and MosMedDataPlus where clinical reports might be longer.

\begin{figure}[!htbp]
    \centering
    \includegraphics[width=1.0\textwidth]{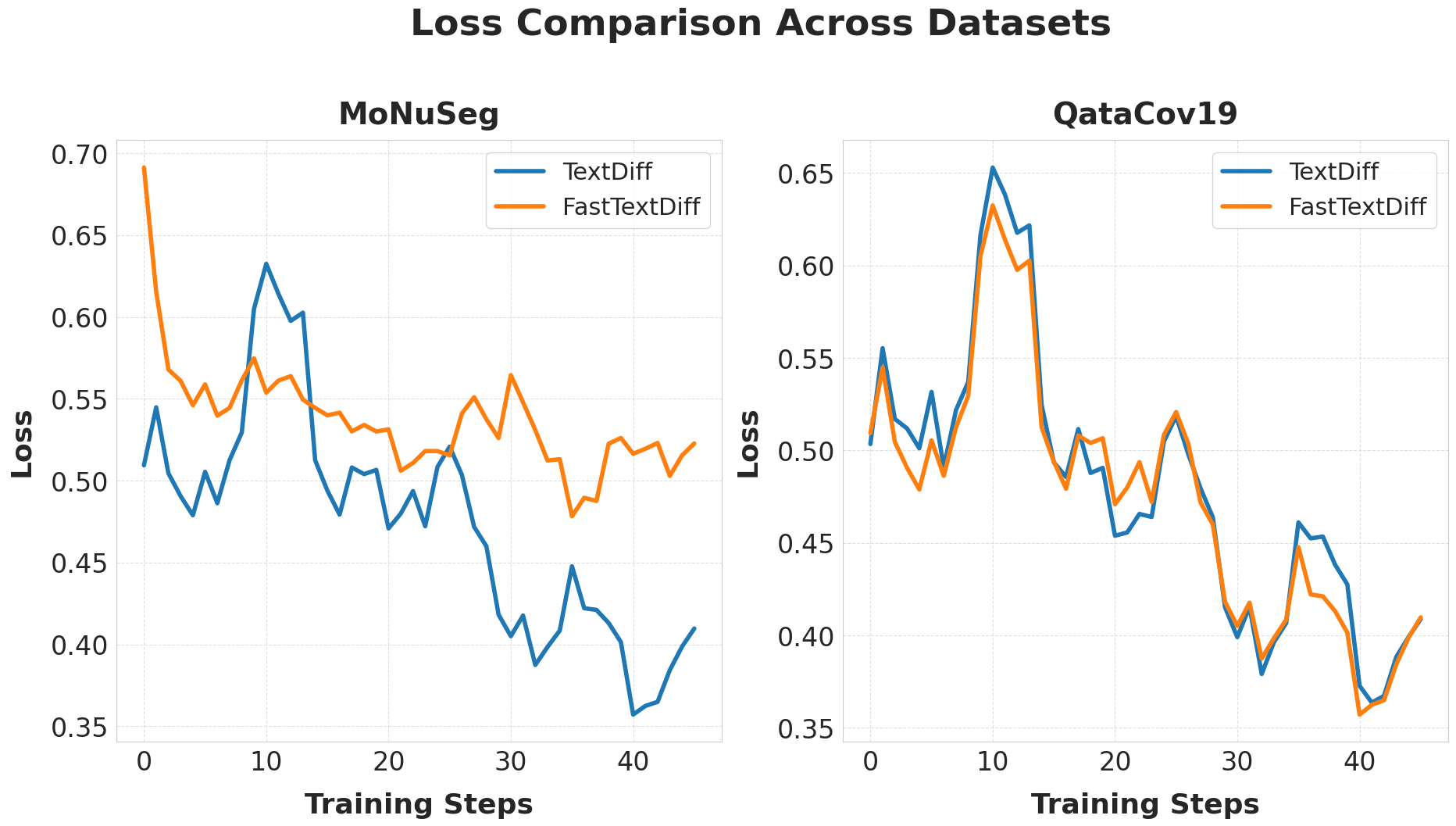}
    \caption{\textbf{Loss Comparison across 2 datasets for TextDiff (baseline BERT) and FastTextDiff (ModernBERT)}. FastTextDiff generally shows faster convergence or reaches a lower loss plateau.}
    \label{fig:loss_comparison} 
\end{figure}

\begin{table}[!htbp]
    \centering
    \caption{\textbf{Performance (Dice/IoU) of Different Text Encoders within the FastTextDiff Framework}. Here mBERT stands for ModernBERT was pre-trained on MIMIC IV for 4 epochs. Numbers 128/512 denote context length. Baseline BERT corresponds to TextDiff. \cite{feng2024enhancing}. Here, MoNu, QaTa, MosMed stand for MoNuSeg, Qata Cov19 and MosMedDataPlus respectively.}
    \label{tab:results}
    \begin{tabular}{
        |>{\raggedright\arraybackslash}p{2.9cm}
        |>{\centering\arraybackslash}p{1.4cm}
        |>{\centering\arraybackslash}p{1.4cm}
        |>{\centering\arraybackslash}p{1.4cm}
        |>{\centering\arraybackslash}p{1.4cm}
        |>{\centering\arraybackslash}p{1.5cm}
        |>{\centering\arraybackslash}p{1.5cm}|}
        \hline
        \textbf{Text Encoder Variant} & MoNu Dice (\%) & MoNu IoU (\%) & QaTa Dice (\%) & QaTa IoU (\%) & MosMed Dice (\%) & MosMed IoU (\%) \\
        \hline
        BERT (w/o MIMIC fine-tune, TextDiff baseline) & 77.50 & 63.41 & 73.54 & 60.96 & 67.20 & 53.64 \\
        \hline
        mBERT (w/o MIMIC fine-tune) & 78.90 & 65.27 & 74.12 & 61.24 & 67.94 & 54.91 \\
        \hline
        mBERT (MIMIC IV fine-tune, 128 context) & 78.71 & 65.00 & 74.29 & 61.47 & 67.88 & 54.24 \\
        \hline
        mBERT (MIMIC IV fine-tune, 512 context) & \textbf{78.72} & \textbf{65.03} & \textbf{74.43} & \textbf{61.75} & 67.39 & 54.36 \\
        \hline
        Clinical BioBERT (MIMIC III fine-tune, TextDiff \cite{feng2024enhancing}) & 78.67 & 64.98 & 71.41 & 59.03 & \textbf{68.99} & \textbf{55.37} \\
        \hline
    \end{tabular}
\end{table}

Table~\ref{tab:sota_comparison} compares FastTextDiff (using ModernBERT MIMIC IV, 512 context) against several state-of-the-art medical image segmentation methods which include UNet \cite{ronneberger2015u}, TransUNet \cite{chen2021transunet}, SwinUNet \cite{cao2021swin}, GLoRIA \cite{huang2021gloria}, LViT \cite{li2022lvit}, and the original TextDiff \cite{feng2024enhancing}. FastTextDiff achieves competitive or superior performance, particularly on MoNuSeg and QaTa-COV19, despite having significantly fewer parameters than transformer-based methods like TransUNet and SwinUNet. Note that FastTextDiff results are reported after 50 epochs, while others are reported after 100 epochs in the cited source \cite{feng2024enhancing}, highlighting its training efficiency.

\begin{table}[!htbp]
    \centering
    \caption{\textbf{Training and Inference Time Comparison (in seconds)}. TextDiff uses Clinical BioBERT \cite{alsentzer2019publicly} fine-tuned on MIMIC-III. FastTextDiff uses ModernBERT (shown as mBERT in table) \cite{modernbert} fine-tuned on MIMIC-IV (4 epochs). Training times are per 50 epochs for segmentation task. ModernBERT Pre-training time is a one-off cost. Here, MoNu, QaTa, MosMed stand for MoNuSeg, Qata Cov19 and MosMedDataPlus respectively.}
    \label{tab:time_results}
    \begin{tabular}{
        |>{\centering\arraybackslash}p{2.1cm}
        |>{\centering\arraybackslash}p{1.6cm}
        |>{\centering\arraybackslash}p{1.1cm}
        |>{\centering\arraybackslash}p{1.3cm}
        |>{\centering\arraybackslash}p{1.4cm}
        |>{\centering\arraybackslash}p{1.2cm}
        |>{\centering\arraybackslash}p{1.2cm}
        |>{\centering\arraybackslash}p{1.4cm}|}
        \hline
        Method & mBERT (Seconds) & \multicolumn{3}{p{3.9cm}|}{\centering Segmentation Training Time} & \multicolumn{3}{p{3.9cm}|}{\centering Inference Time (per image/batch)} \\
        \hline
        & & MoNu & QaTa & MosMed & {MoNu} & QaTa & MosMed \\
        \hline
        TextDiff \cite{feng2024enhancing} & N/A & 664.54 & 3018.23 & 8027.64 & 2.14 & 6.56 & 15.25 \\
        \hline
        FastTextDiff (Ours) & 4388.97 & \textbf{532.17} & \textbf{1891.63} & \textbf{6972.06} & \textbf{1.67} & \textbf{3.97} & \textbf{12.51} \\
        \hline
    \end{tabular}
\end{table}

\subsubsection{Efficiency Analysis.}
Figure~\ref{fig:loss_comparison} illustrates the training dynamics. FastTextDiff generally converges faster (reaching lower loss and higher Dice score earlier) compared to the TextDiff baseline, attributed to the efficiency of ModernBERT \cite{modernbert} and potentially better feature alignment facilitated by its representations.

Table~\ref{tab:time_results} quantifies the training and inference time improvements. FastTextDiff (using ModernBERT) demonstrates significantly reduced training time per 50 epochs compared to TextDiff (using Clinical BioBERT \cite{alsentzer2019publicly}) across all datasets. Inference time, as shown in (Table~\ref{tab:time_results}, is also notably faster. The initial ModernBERT pre-training time (4388.97s reported in Table~\ref{tab:time_results}) is a one-time cost for adapting the language model to the clinical domain (MIMIC IV).

\begin{table}[!htbp]
    \centering
    \caption{\textbf{Comparison with State-of-the-Art Methods}. Results for other methods are cited from \cite{feng2024enhancing} (trained for 100 Epochs). FastTextDiff was trained for 50 epochs. Here, MoNu, QaTa, MosMed stand for MoNuSeg, Qata Cov19 and MosMedDataPlus respectively. Best results for each dataset are shown in bold.}
    \label{tab:sota_comparison}
    \begin{tabular}{
        |>{\raggedright\arraybackslash}p{1.9cm}
        |>{\centering\arraybackslash}p{1.1cm}
        |>{\centering\arraybackslash}p{1.3cm}
        |>{\centering\arraybackslash}p{1.2cm}
        |>{\centering\arraybackslash}p{1.3cm}
        |>{\centering\arraybackslash}p{1.5cm}
        |>{\centering\arraybackslash}p{1.5cm}
        |>{\centering\arraybackslash}p{1.4cm}|}
        \hline
        Method & \# Param. (M) & MoNu Dice (\%) & MoNu IoU (\%) & QaTa Dice (\%) & QaTa IoU (\%) & MosMed Dice (\%) & MosMed IoU (\%) \\
        \hline
        UNet \cite{ronneberger2015u} & 31.04 & 73.92 & 58.98 & 46.08 & 34.16 & 64.60 & 50.73 \\
        \hline
        TransUNet \cite{chen2021transunet} & 93.19 & 73.54 & 58.79 & 70.78 & 59.50 & 71.24 & 58.44 \\
        \hline
        SwinUNet \cite{cao2021swin} & 27.17 & 64.36 & 48.74 & 65.19 & 51.87 & 63.29 & 50.19 \\
        \hline
        GLoRIA \cite{huang2021gloria} & 32.52 & 66.38 & 49.83 & 71.05 & 59.74 & 72.42 & 60.18 \\
        \hline
        LViT \cite{li2022lvit} & 29.72 & 57.95 & 44.13 & 66.43 & 51.71 & 74.57 & 61.33 \\
        \hline
        TextDiff \cite{feng2024enhancing} & 9.68 & 78.67 & 64.98 & 71.41 & 59.03 & \textbf{68.99} & \textbf{55.37} \\
        \hline
        FastTextDiff (Ours) & 9.68 & \textbf{78.72} & \textbf{65.03} & \textbf{74.43} & \textbf{61.75} & 67.39 & 54.36 \\
        \hline
    \end{tabular}
\end{table}

\begin{figure}[!t]
    \centering
    \includegraphics[width=1.0\linewidth]{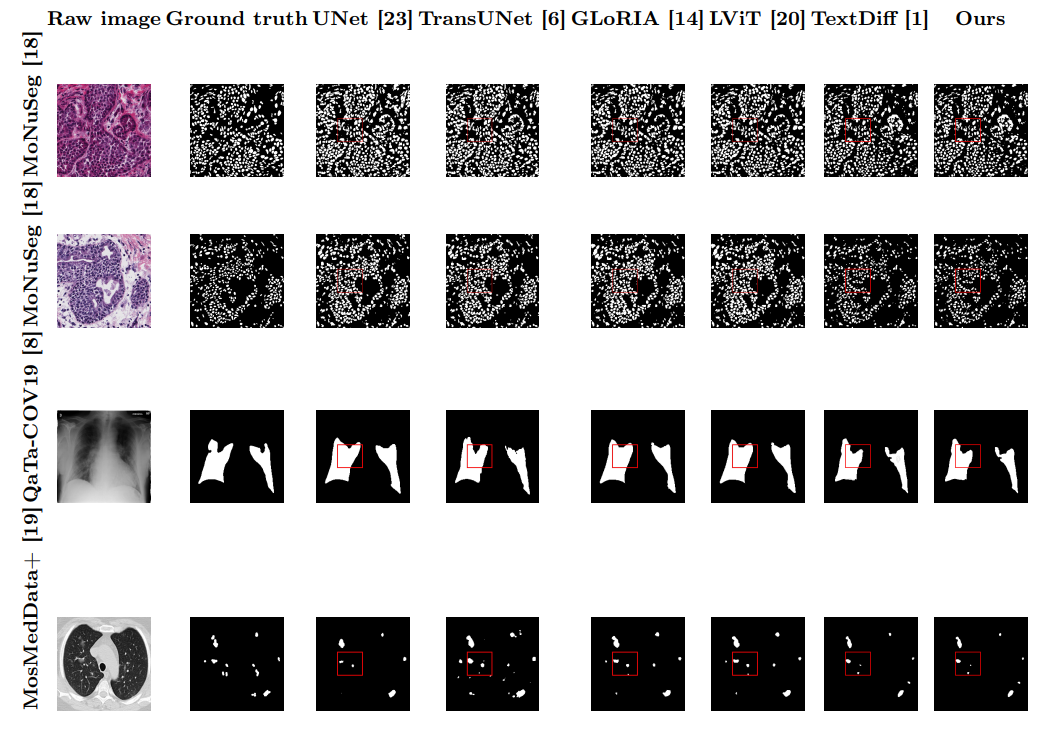}
    \caption{Visual segmentation comparisons on representative images from MoNuSeg \cite{kumar2019multi}, QaTa-COV19 \cite{degerli2022osegnet}, and MosMedData+ \cite{li2021dual}. FastTextDiff often produces segmentations similar to the ground truth compared to other state-of-the-art methods.}
    \label{fig:segmentation_comparison}
\end{figure}

\section{Discussion}

Our experimental results demonstrate that FastTextDiff, by incorporating ModernBERT \cite{modernbert} into a text-conditioned diffusion model framework \cite{feng2024enhancing}, attains both strong segmentation performance and significant efficiency advantages.

\noindent\textbf{Interpretation of Results:} The core hypothesis that a modern, efficient language model like ModernBERT could enhance text-conditioned segmentation is well-supported by our findings. Table~\ref{tab:sota_comparison} shows FastTextDiff achieves Dice/IoU scores that are competitive with or better than the TextDiff baseline (using Clinical BioBERT \cite{alsentzer2019publicly}) and other state-of-the-art techniques across multiple datasets. Notably, it outperforms established methods like UNet \cite{ronneberger2015u} and even sophisticated transformer architectures like TransUNet \cite{chen2021transunet} and SwinUNet \cite{cao2021swin} on MoNuSeg \cite{kumar2019multi} and QaTa-COV19 \cite{degerli2022osegnet}, despite having significantly fewer parameters (9.68M vs 93.19M for TransUNet). This suggests that the semantic guidance provided by the text, efficiently captured by ModernBERT and fused via cross-modal attention, allows the relatively lightweight diffusion backbone to achieve high accuracy, potentially reducing the need for extremely complex visual-only feature extractors. The ablation study in Table~\ref{tab:results} further confirms the benefit of ModernBERT over a standard BERT and shows that MIMIC pre-training and longer context lengths (512) generally yield the best results, highlighting its ability to leverage more extensive clinical context. The qualitative results in Figure~\ref{fig:segmentation_comparison} visually corroborate these quantitative findings.

\noindent\textbf{Efficiency Gains:} A key motivation for adopting ModernBERT was efficiency. Table~\ref{tab:time_results} and Figures ~\ref{fig:loss_comparison}, and (Table~\ref{tab:time_results} clearly demonstrate these advantages. FastTextDiff exhibits significantly faster training convergence and reduced training time per epoch compared to the TextDiff baseline across all datasets. Inference speed is also considerably improved. These gains stem from ModernBERT's architecture: Flash Attention 2 optimizes attention computation speed and memory usage, while its alternating sparse/dense attention patterns reduce the computational load, especially beneficial for potentially long clinical text inputs. This efficiency makes FastTextDiff more practical for training on large datasets and for potential deployment in time-sensitive clinical workflows.

\noindent\textbf{Comparison to State-of-the-Art:} FastTextDiff positions favorably against other approaches (Table~\ref{tab:sota_comparison}). Compared to vision-only models (UNet \cite{ronneberger2015u}, TransUNet \cite{chen2021transunet}, SwinUNet \cite{cao2021swin}), its use of textual information contributes to its strong performance, especially where visual cues might be ambiguous but text provides disambiguation. Compared to other vision-language models (GLoRIA \cite{huang2021gloria}, LViT \cite{li2022lvit}), FastTextDiff uniquely employs a diffusion model backbone, potentially benefiting from the rich, hierarchical features learned during the diffusion process \cite{baranchuk2021label,feng2023diverse}. Its parameter efficiency compared to large transformer models is also a significant advantage. While TextDiff \cite{feng2024enhancing} introduced the core concept, FastTextDiff significantly improves upon it through the use of a more advanced and efficient language model.

\noindent\textbf{Limitations and Future Work:} While FastTextDiff shows promise, some limitations exist. The performance on MosMedData+ was slightly lower than the original TextDiff, potentially due to differences in the fine-tuning data (MIMIC IV vs MIMIC III) or the specific characteristics of that dataset requiring features better captured by Clinical BioBERT. Further investigation into optimal pre-training and fine-tuning strategies for ModernBERT in this context is warranted. Future research could explore different diffusion model backbones or more sophisticated cross-modal fusion techniques beyond standard attention. Applying the framework to weakly-supervised or semi-supervised settings \cite{yu2019uncertainty}, where only text descriptions and unlabeled images are available, could significantly broaden its impact and address the annotation bottleneck more directly. Finally, enhancing the interpretability of how text guides segmentation would increase clinical trust and utility.

\section{Conclusion}

In this paper, we addressed the challenge of label-efficient medical image segmentation by proposing FastTextDiff, a novel framework that successfully incorporates textual information using a modern language model within a diffusion-based segmentation pipeline. By leveraging ModernBERT as an efficient and powerful text encoder, coupled with cross-modal attention mechanisms, our method successfully guides the segmentation process using clinical text descriptions.

Our extensive experiments on three distinct medical imaging datasets: MoNuSeg \cite{kumar2019multi}, QaTa-COV19 \cite{degerli2022osegnet}, and MosMedData+ \cite{li2021dual} demonstrated that FastTextDiff achieves competitive and state-of-the-art segmentation accuracy compared to previous methods and a baseline text-conditioned model. Importantly, we demonstrated notable gains in computational performance, with ModernBERT's architectural optimizations responsible for noticeably reduced training and inference times (Table~\ref{tab:time_results}, Figures~\ref{fig:loss_comparison}-(Table~\ref{tab:sota_comparison}). This work validates ModernBERT as a potent and practical choice for text encoding in multimodal medical imaging tasks and underscores the potential of combining diffusion models with natural language processing for creating powerful, yet more label-efficient, analysis tools.

By making our model publicly available, we intend to encourage research into multimodal learning approaches for medicatableing in the development of tools that can more effectively leverage the abundance of data found in clinical data repositories to enhance patient care.

\begin{credits}
\subsubsection{\ackname}  Venkata Siddharth Dhara thanks IHub-Data, IIIT Hyderabad for extending research fellowship.

\subsubsection{\discintname}
Both authors declare that there are no competing interests.

\end{credits}

\end{document}